\documentclass[12pt]{l4dc2021}

\title[Nested Mixture of Experts]{Nested Mixture of Experts: Cooperative and Competitive Learning of Hybrid Dynamical System}
\usepackage{times}
\usepackage{wrapfig}
\usepackage{mathtools}
\usepackage{hyperref}
\author{%
 \Name{Junhyeok Ahn} \Email{junhyeokahn91@utexas.edu}\\
 \addr Department of Mechanical Engineering, The University of Texas at Austin, USA
 \AND
 \Name{Luis Sentis} \Email{lsentis@austin.utexas.edu}\\
 \addr Department of Aerospace Engineering and Engineering Mechanics, The University of Texas at Austin, USA%
}

\begin{document}

\maketitle

\begin{abstract}%
Model-based reinforcement learning (MBRL) algorithms can attain significant sample efficiency but require an appropriate network structure to represent system dynamics. Current approaches include white-box modeling using analytic parameterizations and black-box modeling using deep neural networks. However, both can suffer from a bias-variance trade-off in the learning process, and neither provides a structured method for injecting domain knowledge into the network. As an alternative, gray-box modeling leverages prior knowledge in neural network training but only for simple systems. In this paper, we devise a nested mixture of experts (NMOE) for representing and learning hybrid dynamical systems. An NMOE combines both white-box and black-box models while optimizing bias-variance trade-off. Moreover, an NMOE provides a structured method for incorporating various types of prior knowledge by training the associative experts cooperatively or competitively. The prior knowledge includes information on robots' physical contacts with the environments as well as their kinematic and dynamic properties. In this paper, we demonstrate how to incorporate prior knowledge into our NMOE in various continuous control domains, including hybrid dynamical systems. We also show the effectiveness of our method in terms of data-efficiency, generalization to unseen data, and bias-variance trade-off. Finally, we evaluate our NMOE using an MBRL setup, where the model is integrated with a model-based controller and trained online.
\end{abstract}


\section{Introduction}
Reinforcement learning algorithms provide an automated framework for decision making by maximizing an objective function and learning a control policy. However, model-free reinforcement learning algorithms require a vast amount of data for learning policies, which often limits their application to simulated domains (\cite{td3, ppo}). An alternative method for reducing sample complexity is to explore model-based reinforcement learning (MBRL) methods, which involve obtaining a predictive model of the world and then using that model to make decisions (\cite{pets, mbmf, gps}).

In MBRL, model representation is one of the most critical design choices and determines the performance and efficiency of the algorithm. One option is to treat dynamics as a function where machine learning is capable of fitting. For example, \cite{pets,mbmf,multistep_sysid} deployed deep neural networks and took a data-driven, black-box approach to approximate dynamics. On the other hand, some works utilize a white-box approach that exploits a priori domain knowledge, make assumptions about kinematic and dynamic properties, and leave a few parameters for identification. For instance, \cite{inertia_id,dracop2} conducted classical system identification to learn inertial parameters, and \cite{brl} integrated model learning and planning using Bayesian reinforcement learning. The black-box approach relies on the network's representational power but often requires tremendously large amounts of training data to achieve generalization and suffers from model variance. The assumptions in the white-box approach may not capture hard-to-model effects, leading to model bias, although they enable training with much less data.

Other studies leverage a gray-box method to mitigate the limitations. \cite{delan2, delan1} and \cite{rnea} proposed a network topology that encoded the Euler-Lagrange and Newton-Euler equation, respectively, and demonstrated that their model is trained efficiently and extrapolates effectively to unseen samples. \cite{hamil1,hamil2} presented a dynamics model that conformed to the Hamiltonian equation and demonstrated that their model conserves a quantity analogous to the total energy. However, they had to predefine the dynamics equation structure in advance and train the model under the prescribed model capacity, which could still cause a model bias. \cite{residual1,residual2,residual3} fitted residual dynamics with a Gaussian process based upon a prescribed analytic model while accounting for hard-to-model effects and reduced model bias. The aforementioned gray-box approaches are all limited to representing simple mechanical systems, such as a pendulum or robotic arms, where a single dynamic equation prescribes the behaviors over the state space. For hybrid dynamical systems that may include intricate prior knowledge, such as physical contacts with the environment, multiple dynamic models need to be considered instead of a single model.

Thus, the purpose of this paper is to study a gray-box approach that can be used for efficient learning of complex robotic systems, including hybrid systems. To that end, we propose a network structure, dubbed a nested mixture of experts (NMOE), to represent and learn dynamic models. Using this novel network structure, we propose a general method to incorporate various types of physical insights by training local experts to be cooperative or competitive. We demonstrate the effectiveness of our NMOE with useful metrics in various continuous control domains. We further evaluate our NMOE using an MBRL setup, where the model is integrated with a model-based controller and trained online.

\section{Preliminaries}
\begin{wrapfigure}{r}{0.3\textwidth}
  \centering
    \includegraphics[width=0.3\textwidth]{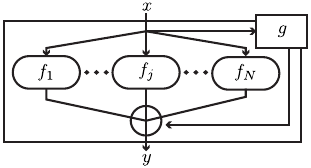}
    \vspace{-6mm}
  \caption{Mixture of experts.}
  \vspace{-2mm}
  \label{fig:moe}
\end{wrapfigure}
A mixture of experts (MOE) is a network composed of many separate feedforward sub-networks and ensembles outputs of the sub-networks to improve model performance in the supervised learning procedure (\cite{moe}). It includes a gating network that decides what combination of the experts should be used for each data input. Note that the gating network is also feedforward and typically receives the same input as the expert networks. Figure~\ref{fig:moe} shows an example of an MOE with $N$ local experts and a gating network $g$ that computes outputs $y_i = \sum_{k=1}^{N} g_{k}(x_i)f_{k}(x_i)$, where $\sum_{k=1}^{N}g_{k}(x_i) = 1$. Here, $x_i$ and $y_i$ are the input and the output of the MOE given a dataset made of pairs $(x_i,y_{\text{target},i})$. The weight $g_{k}$ corresponds to the $k$th element of $g$ and indicates the probability (or so called responsibility) for the expert $f_k$. \cite{moe} showed that each of the experts is trained to either cooperate or compete to produce the output based on a loss function. For example, a loss function with the expression $\sum_{i=1}^{D} \lVert y_{\text{target,i}} - y_i \rVert ^2$, or equivalently $\sum_{i=1}^{D} \lVert y_{\text{target,i}} - \sum_{k=1}^{N} g_{k}(x_i)f_{k}(x_i) \rVert ^2$, makes the experts contribute to the output linearly, where $D$ is the size of the dataset. With this loss function, each expert is trained to produce an output that eliminates the residual errors left by the other experts. When the local experts cooperate to generate the output, we call this a cooperative MOE. Conversely, another loss function $\sum_{i=1}^{D} \sum_{k=1}^{N} g_k(x_i) \lVert y_{\text{target},i} - f_{k}(x_i) \rVert ^2$ makes each local expert produce the whole output rather than a residual, and respond only to a subset of the dataset. If we train the network with gradient descent using this loss function, the MOE tends to devote a single expert to each training case. As a result, each expert is not directly affected by the weights of other experts, which can be useful when we know in advance that a set of training data is naturally classified into particular expert categories. When the local experts compete to generate the output, we call this a competitive MOE.

\section{Nested Mixture of Experts}

\subsection{Network Structure}
\label{sec:ccm}
As depicted in Figure~\ref{fig:ccm_structure}, an NMOE has a nested structure of an MOE. The top layer (pink box) forms a competitive MOE with $M$ local experts (blue boxes) and the gating network (purple box). As a nested structure, each of the local experts (blue boxes) forms a cooperative MOE containing two local experts (ellipsoids) and a gating network (green box)\footnote{Here, we only consider two local experts in the bottom layer, although an MOE can have any number of local experts.}. The $i$th local expert in the top layer ($i$th blue box), or equivalently $i$th cooperative MOE, computes an output as follows:
\begin{figure*}[t]
    \centering
    \includegraphics[width=0.7\linewidth]{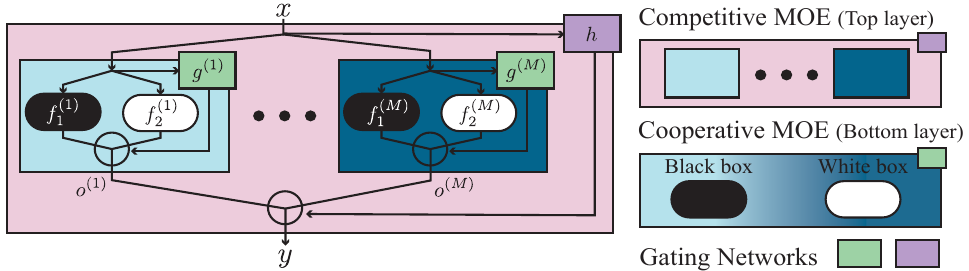}
    \caption{NMOE structure: The top layer forms a competitive MOE, and each of the experts on it are cooperative MOEs.}
    \label{fig:ccm_structure}
\end{figure*}
\begin{equation}
\begin{split}
     o^{(i)}(x) &= \sum^{2}_{j=1}g^{(i)}_j(x)f^{(i)}_j(x), \\
     g^{(i)}(x) &= \text{softmax}(W^{(i)}_g x).
\end{split}
\end{equation}
Here, $x$ is the input, $o^{(i)}(x)$ is the output of the $i$th cooperative MOE, and $f^{(i)}_j(x)$ is the output of the $j$th expert in the $i$th cooperative MOE. $g^{(i)}_j(x)$, the $j$th element of $g^{(i)}(x)$, indicates the probability (or responsibility) for the expert and $W^{(i)}_g$ is a trainable weight. A competitive MOE (pink box) combines the outputs of $M$ cooperative MOEs (blue boxes) as follows:
\begin{equation}
    \begin{split}
        y &= \sum^{M}_{i=1} h_i(x)o^{(i)}(x), \\
        h(x) &= \text{softmax}(W_h x),
    \end{split}
\end{equation}
where $y$ is the final output of the NMOE, and $h_i(x)$ is the $i$th element of $h(x)$, indicating the probability (or responsibility) for the output $o^{(i)}$ with a trainable weight $W_h$. Finally, we define a loss function as follows:
\begin{equation*}
\begin{split}
\label{eq:loss}
    \text{loss}(x,y_{\text{target}}) = \frac{1}{M}\sum^{M}_{i=1} h_i(x) \left\lVert y_{\text{target}} - o^{(i)}(x) \right\rVert^2
    = \frac{1}{M}\sum^{M}_{i=1} h_i(x) \left\lVert y_{\text{target}} - \sum^{2}_{j=1}g^{(i)}_j(x)f^{(i)}_j(x) \right\rVert^2,
\end{split}
\end{equation*}
so that the $M$ local experts (blue boxes) in the top layer are trained to compete with each other, whereas the two local experts (ellipsoids) in the bottom layer are trained to cooperate. Due to this loss function, each of the cooperative MOEs (blue boxes) produces the whole of the output and correlates with a subset of the dataset that is determined by their responsibility (i.e., the output of the gating network $h$). In contrast, the two local experts (ellipsoids) inside the cooperative MOEs contribute to the output linearly, which is determined by their responsibility (i.e., the output of the gating network $g$). In summary, given the input $x$, the competitive MOE (pink box) assigns the input to a single cooperative MOE (blue box). The chosen cooperative MOE combines the outputs from its local experts (ellipsoids) to generate the final output $y$. Note that the input $x$ corresponds to a state and action pair, and the output $y$ corresponds to a next-state prediction when it comes to dynamic model representation.

\subsection{Incorporating Prior Knowledge}
In this section, we discuss how to constitute an NMOE to incorporate prior knowledge. We consider two different types of prior knowledge. The first type includes information on physical contacts between a system and an environment. In the case of a hybrid dynamical system, we know in advance that the input dataset is naturally divided into subsets based on the contact modes, and each of the subsets is governed by a different physics law. This leads us to set $M$, the number of experts in the competitive MOE, to be the number of contact modes. The gating network (purple box) divides input data into $M$ subsets and assigns each of them to a local expert associated with the contact mode. A competitive MOE can significantly reduce the model complexity compared to using a single network since each local expert is associated with a subset of the dataset.

The second type of prior knowledge corresponds to system equations describing the robot's behavior with assumptions on rigid body kinematics and dynamics. We then compose a cooperative MOE (blue box) combining the system equations (i.e., the white-box model) and a deep neural network (i.e., black-box model) as local experts (ellipsoids). The white-box model provides a reasonable initial representation of the model and aids in achieving quick convergence of the learning process with low variance. In contrast, the black-box model eliminates the residual errors left by the hard-to-model effects and reduces the bias. The gating network (green box) computes a probability (or responsibility) based on the reliability of the prior knowledge.

Putting these together, we consider a structured method to incorporate prior knowledge into an NMOE. We first estimate the type of robot's contact modes, $M$. We then define a competitive MOE (pink box) with $M$ local experts, each of them being cooperative MOEs (blue boxes). Each of the cooperative MOEs (blue boxes) includes a white-box (white ellipsoid) and a black-box model (black ellipsoid), where the white-box model is a differential equation of motion and the black-box model is a deep neural network.
%===============================================================================

\section{Experimental Results}
\label{sec:exp}
\begin{figure}[t]
    \centering
    \includegraphics[width=0.8\linewidth]{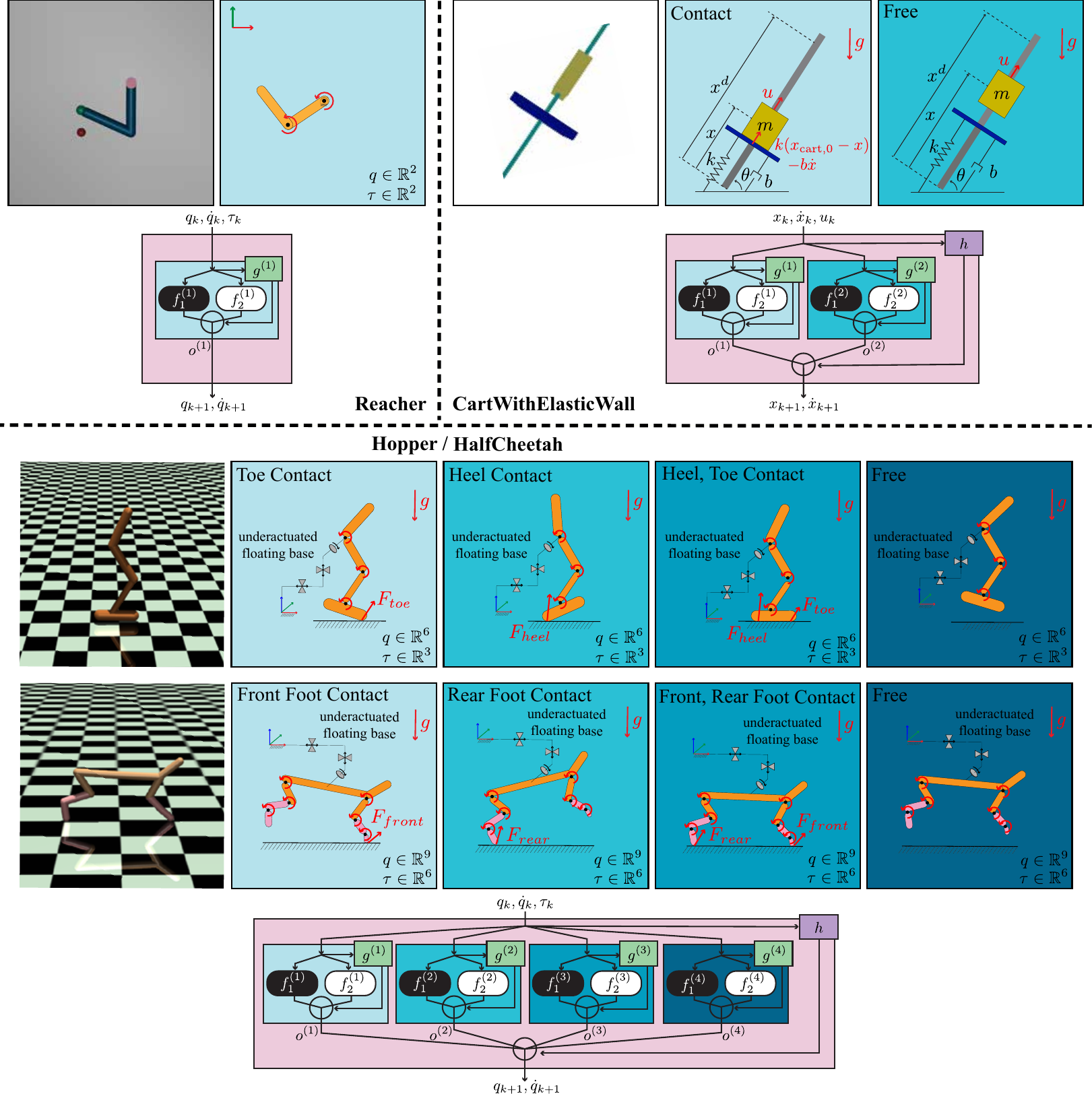}
    \caption{Illustration of physical contact modes and NMOE structures for various domains.}
    \label{fig:exp_domains}
\end{figure}
In this study, we use four different continuous control domains (Reacher, CartWithElasticWall, Hopper, and HalfCheetah) and address the following questions: How do we construct an NMOE for each domain based on prior knowledge? Does our NMOE framework outperform related baseline approaches in dynamic model learning? Does the top layer's gating network (purple box) successfully assign input data to a corresponding contact mode after convergence? Does the bottom layer's gating network (green box) properly combine white-box and black-box models based on the accuracy of the prior knowledge? In an MBRL setup, can we train NMOE online and provide reliable trajectory predictions to the model-based controller? Parameters used in dynamics model representations, model learning, and model-based controller for each domain can be found in Appendix~\ref{sec:app_a}
% arxiv

\subsection{NMOE Design and Baseline Models}
To design an NMOE for each experimental domain, we estimate the number of contact modes of the robot and formulate a competitive MOE. The type of contact modes are illustrated in Figure~\ref{fig:exp_domains}. For each contact mode, we derive a white-box model based on first principles using estimations on the robot kinematics and inertial parameters. Note that these parameters are initialized within a $50$ percent margin of error. We derive the equations of motion for the white-box models using the free body diagrams illustrated in Figure~\ref{fig:exp_domains}. We have developed a new rigid body dynamics library using Tensorflow to encode the white-box models (\cite{tf_rbdl}). In addition, we build the black-box models with a multi-layer perceptron denoted by \texttt{MLP}. The equations in this section use the nomenclature shown in Figure~\ref{fig:exp_domains}.

\textbf{Reacher:} Reacher has a state $[q; \dot{q}] \in \mathbb{R}^{4}$ and an action $\tau \in \mathbb{R}^2$ representing joint positions, velocities, and torques. The NMOE for the Reacher uses a single cooperative MOE since there is a single contact mode (a fixed base manipulator). It computes the output with the expression:
\begin{equation}
\begin{split}
    \begin{bmatrix} q_{k+1} \\ \dot{q}_{k+1} \end{bmatrix} = \sum_{j=1}^{2} g_{j}^{(1)} f^{(1)}_j(q_k, \dot{q}_k, \tau_k)
    = g^{(1)}_1 \underbrace{\texttt{MLP}_{\theta_1}(q_k, \dot{q}_k, \tau_k)}_{\text{black-box}} + g^{(1)}_2 \underbrace{\begin{bmatrix}q_k + \dot{q}_k dt \\ \dot{q}_{k} + \left( A^{-1}(\tau_k-b-g) \right) dt\end{bmatrix}}_{\text{white-box}},
\end{split}
\label{eq:reacher}
\end{equation}
where $A \in \mathbb{R}^{2 \times 2}$, $b \in \mathbb{R}^{2}$, and $g \in \mathbb{R}^{2}$ represent the mass matrix, centrifugal / coriolis, and gravity vectors, respectively. Trainable variables in the NMOE include the mass and the length of each link as well as the parameters in the \texttt{MLP} and the gating networks.

\textbf{CartWithElasticWall:} This system has a single degree of freedom cart sliding along the inclined pin. The cart has to interact with the elastic wall down below and accumulate energy to reach a desired height accounting for actuator limits. The cart has a state $[x; \dot{x}] \in \mathbb{R}^{2}$ and an input $u\in \mathbb{R}$ representing the cart's position, velocity, and force. The NMOE includes two cooperative MOEs since there are two possible contact modes: an interacting mode with the elastic wall and a free-floating mode when there is no wall contact. The output is computed as 
\begin{align}
    &\begin{bmatrix} x_{k+1} \\ \dot{x}_{k+1} \end{bmatrix} = \sum_{i=1}^{2}h_i \left(\sum_{j=1}^{2} g_{j}^{(i)} f^{(i)}_j(x_k, \dot{x}_k, u_k)\right) \nonumber \\
    &= h_1\overbrace{\bigg(g^{(1)}_1 \underbrace{\texttt{MLP}_{\theta_1}(x_k, \dot{x}_k, u_k)}_{\text{black-box}} + g^{(1)}_2 \underbrace{\begin{bmatrix}x_k + \dot{x}_k dt \\ \dot{x}_{k} + \frac{1}{m}\left( u_k+k(x_{\text{cart,0}}-x_k)-b\dot{x}_k-mg\sin \theta \right) dt\end{bmatrix}}_{\text{white-box}}\bigg)}^{\text{1st cooperative MOE: Contact}} \nonumber \\ 
    & + h_2\overbrace{\bigg(g^{(2)}_1 \underbrace{\texttt{MLP}_{\theta_2}(x_k, \dot{x}_k, u_k)}_{\text{black-box}} + g^{(2)}_2 \underbrace{\begin{bmatrix}x_k + \dot{x}_k dt \\ \dot{x}_{k} + \frac{1}{m}\left( u_k-mg\sin \theta \right) dt\end{bmatrix}}_{\text{white-box}}\bigg)}^{\text{2nd cooperative MOE: Free}},
    \label{eq:cart}
\end{align}
where $m$ and $x_{\text{cart,0}}$ are the mass and initial position of the cart, $k$ and $b$ are the spring constant and the damping of the elastic wall, respectively, $g$ is the gravitational acceleration, $\theta$ is the angle of the rail, and $dt$ is the integration step. Training variables in the NMOE include $m$, $k$, $b$, $\theta$, the parameters in the \texttt{MLP}s, and the gating networks.

\textbf{Hopper:} The Hopper has a state $[q; \dot{q}]\in\mathbb{R}^{12}$ and an input $\tau \in \mathbb{R}^{3}$ representing joint positions, velocities, and torques. The NMOE includes four cooperative MOEs since there are four possible contact modes: toe contact mode, heel contact mode, heel-toe contact mode, and contact free mode. The output is computed as 
\begin{align}
    \label{eq:hopper_ccm}
    &\begin{bmatrix} q_{k+1} \\ \dot{q}_{k+1} \end{bmatrix} = \sum_{i=1}^{4}h_i \left(\sum_{j=1}^{2} g_{j}^{(i)} f^{(i)}_j(q_k, \dot{q}_k, \tau_k)\right) \nonumber \\
    &= h_1\overbrace{\bigg(g^{(1)}_1 \underbrace{\texttt{MLP}_{\theta_1}(q_k, \dot{q}_k, \tau_k)}_{\text{black-box}} + g^{(1)}_2 \underbrace{\begin{bmatrix}q_k + \dot{q}_k dt \\ \dot{q}_{k} + A^{-1}(N_{\text{T}}^{\top}(S^{\top}\tau_k-b-g)-J_{\text{T}}^{\top}\Lambda_{\text{T}}\dot{J}_{\text{T}}\dot{q}) dt\end{bmatrix}}_{\text{white-box}}\bigg)}^{\text{1st cooperative MOE: T(oe) Contact}} \nonumber \\
    & + h_2\overbrace{\bigg(g^{(2)}_1 \underbrace{\texttt{MLP}_{\theta_2}(q_k, \dot{q}_k, \tau_k)}_{\text{black-box}} + g^{(2)}_2 \underbrace{\begin{bmatrix}q_k + \dot{q}_k dt \\ \dot{q}_{k} + A^{-1}(N_{\text{H}}^{\top}(S^{\top}\tau_k-b-g)-J_{\text{H}}^{\top}\Lambda_{\text{H}}\dot{J}_{\text{H}}\dot{q}) dt\end{bmatrix}}_{\text{white-box}}\bigg)}^{\text{2nd cooperative MOE: H(eel) Contact}} \nonumber \\
    & + h_3\overbrace{\bigg(g^{(3)}_1 \underbrace{\texttt{MLP}_{\theta_3}(q_k, \dot{q}_k, \tau_k)}_{\text{black-box}} + g^{(3)}_2 \underbrace{\begin{bmatrix}q_k + \dot{q}_k dt \\ \dot{q}_{k} + A^{-1}(N_{\text{HT}}^{\top}(S^{\top}\tau_k-b-g)-J_{\text{HT}}^{\top}\Lambda_{\text{HT}}\dot{J}_{\text{HT}}\dot{q} ) dt\end{bmatrix}}_{\text{white-box}}\bigg)}^{\text{3rd cooperative MOE: H(eel), T(oe) Contact}} \nonumber \\
    & + h_4\overbrace{\bigg(g^{(4)}_1 \underbrace{\texttt{MLP}_{\theta_4}(q_k, \dot{q}_k, \tau_k)}_{\text{black-box}} + g^{(4)}_2 \underbrace{\begin{bmatrix}q_k + \dot{q}_k dt \\ \dot{q}_{k} + A^{-1}(S^\top \tau_k - b - g) dt\end{bmatrix}}_{\text{white-box}}\bigg)}^{\text{4th cooperative MOE: Contact free}},
\end{align}
where $A \in \mathbb{R}^{6 \times 6}$, $b \in \mathbb{R}^{6}$, and $g \in \mathbb{R}^{6}$ represent the mass matrix, and centrifugal / coriolis, and gravity vectors, respectively. $S \in \mathbb{R}^{3 \times 6}$ is the selection matrix that maps joint space to actuator space. $J_{\text{T}}$ and $J_{\text{H}}$ are both in $\mathbb{R}^{3\times6}$, denoting the jacobian matrices associated with the toe and heel contact frames with respect to the world. $J_{\text{HT}} \in \mathbb{R}^{6 \times 6}$ is a matrix vertically concatenating $J_{\text{H}}$ and $J_{\text{T}}$. $\Lambda_{*} = (J_{*}A^{\top}J_{*}^{\top})^{\top}$ is a mass matrix in frame $*$, $\overline{J}_{*}=A^{-1}J_{*}^{\top}\Lambda_{*}$ is a dynamically consistent pseudo-inversion of $J_{*}$, and $N_{*} = I - \overline{J}_{*}J_{*}$ is a null-space projection operator in frame $*$, where $* \in \{\text{T},\text{H},\text{HT}\}$. Trainable variables in the NMOE include the mass of each link as well as the parameters in the \texttt{MLP} and the gating networks. Readers can refer to Appendix \ref{sec:app_b} for more details about white-box modeling.
%% arxiv

\textbf{HalfCheetah:} The HalfCheetah has a state $[q;\dot{q}]\in\mathbb{R}^{18}$ and an input $\tau \in \mathbb{R}^{6}$ representing joint positions, velocities, and torques. The NMOE includes four cooperative MOEs since there are four possible contacts with the ground: front-foot contact mode, rear-foot contact mode, front- and rear-foot contact mode, and contact free mode. The structure and equations of the NMOE are identical to Eq.~\eqref{eq:hopper_ccm}, with a few modifications (i.e., the dimensions of the kinematic and dynamic operators and the contact frames).

We compare our NMOE models to the following baseline approaches:
\begin{itemize}
\item {\bf Single black-box model (BB)}, which involves a single large size neural network implemented using an \texttt{MLP}.

\item {\bf Mixture of black-box models (MBB)}, being equivalent to our NMOE if the cooperative MOEs include only black-box models. In other words, it is identical to our NMOE when the white-box models are ignored (i.e., $g^{(i)}=[1,~0],~f^{(i)}_2 = 0$). Note that in the case of the Reacher, the BB model and the MBB are equivalent since it only has one mode (i.e., $M=1$).

\item{\bf Mixture of white-box models (MWB)}, which is equivalent to our NMOE when the cooperative MOEs has only white-box models. In other words, it is identical to our NMOE when the black-box models are ignored (i.e., $g^{(i)}=[0,~1],~f^{(i)}_1 = 0$).

\end{itemize}

\subsection{Performance Analysis}
We investigate performance of our NMOE models in terms of data-efficiency, generalization to unseen data, and bias-variance trade-off. For all experimental domains, we uniformly sample state, input, and next-state tuples to build training and test datasets. We then train the NMOE and the baseline models and measure a root mean squared error using the test set. We train the models with various size of training sets ranging from $2^8$ to $2^{12}$ sizes and evaluate them using a test set with $2^{12}$ tuples.
\begin{figure*}[t]
    \centering
    \includegraphics[width=1.0\linewidth]{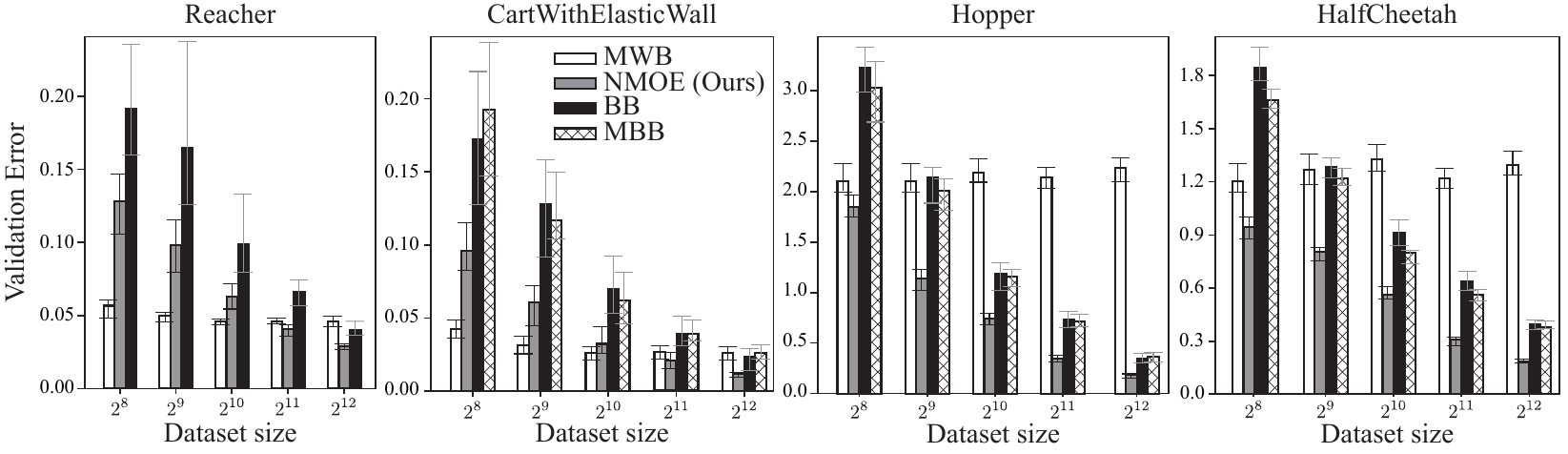}
    \caption{Validation error obtained using our NMOE setup and compared to baseline models for various training data sizes. We plot the median values with interquartile ranges of five runs.}
    \label{fig:ccn}
\end{figure*}
Figure~\ref{fig:ccn} shows the validation error using our NMOE compared to the baseline models demonstrating our NMOE's ability to generalize from limited data. For the cases involving the Hopper and HalfCheetah systems, the MWB model suffers from model bias due to hard-to-model effects and low model capacity, resulting in large generalization errors. The BB and the MBB models show worse data-efficiency and higher model variance in the learning procedure than the other models. We therefore conclude that our NMOE outperforms the baseline models for learning various complex dynamical systems in terms of data-efficiency, generalization, and bias-variance trade-off.

\subsection{Gating Network Evaluation}
\begin{wrapfigure}{r}{0.6\textwidth}
  \centering
    \includegraphics[width=0.6\textwidth]{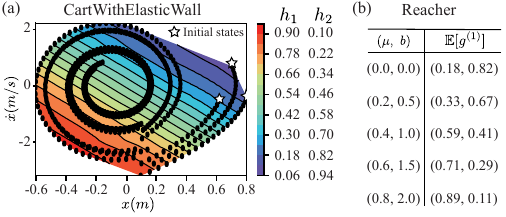}
    \vspace{-6mm}
  \caption{Gating network output after convergence.}
  \vspace{-2mm}
  \label{fig:gating_network}
\end{wrapfigure}

We study how the gating network in the top layer, $h$, in our NMOE assigns input data to local experts associated with the contact modes. Let's consider the NMOE for the CartWithElasticWall domain. After the training converges, we collect trajectories from three randomly sampled initial states with zero inputs and evaluate them. Figure~\ref{fig:gating_network}a shows the outputs of the gating network displayed in the phase plot. Note that $h_1$ and $h_2$ represent the responsibility to the first and second contact modes which correspond the contact and contact-free modes, respectively, as in Eq.~\eqref{eq:cart}. The phase plot demonstrates that the NMOE assigns inputs which have negative displacement and negative velocity, where the cart is most likely contacting the elastic wall, to the first cooperative MOE. Similarly, the NMOE assigns inputs to the second cooperative MOE when the cart is in contact-free mode.

We further demonstrate how the NMOE optimizes the gating network in the bottom layer, $g$, depending on the accuracy of the prior knowledge. In the Reacher domain, we prepare multiple training datasets uniformly sampled from different simulation setups. For example, we set different joint friction and damping coefficients, $\mu$ and $b$, for each dataset. Note that the accuracy of the white-box model in the NMOE decreases as the simulation is setup with large friction and damping since the white-box model is derived under the assumption of zero-friction and zero-damping. Figure~\ref{fig:gating_network}b summarizes the average of the responsibility evaluated over the state-space (i.e., $\mathbb{E}\left[g^{(1)}(q, \dot{q}, \tau)\right]$) for each training case after convergence. Note that the first and the second element of vector $g$ corresponds to the black-box and the white-box model, respectively, as in Eq.~\eqref{eq:reacher}. It shows that the NMOE relies more on the white-box model to generate the output when the prior knowledge is more reliable.

\subsection{Reliability of Model-based Control}
We evaluate our NMOE using an MBRL setup where the network is linked to a model-based controller and is trained online. Here, we devise a model predictive path integral controller (\cite{mppi}) using our NMOE model and compared with other baseline models. The reward function for each experimental domain is described in Appendix \ref{sec:app_c}. % arxiv
\begin{figure*}[t]
    \centering
    \includegraphics[width=1.0\linewidth]{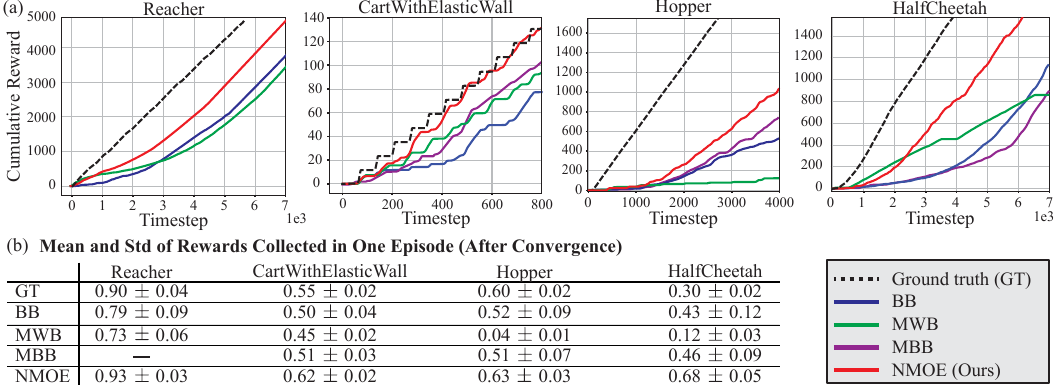}
    \caption{(a) illustrates the average cumulative rewards during the training. (b) represents the final performance of the model-based controllers. We run five experiments with random seeds.}
    \label{fig:result}
\end{figure*}
Figure~\ref{fig:result} summarizes the performance of the model-based controllers. It includes the cumulative rewards in the training procedure and the final performance of the controllers after convergence. The final performance is measured with the mean and std of rewards collected in one episode after the models converge. Note that our NMOE achieves a similar cumulative reward gradient to the ground truth model at the earliest timestep compared to the baseline methods. The MWB model provides a reliable model capacity for the Reacher and the CartWithElasticWall but fails to represent the Hopper and HalfCheetah. The BB model and the MBB model require much more data to converge and show higher variance in the final performance compared to our NMOE.

\section{Conclusion}
\label{sec:conclusion}
In this study, we introduce a novel network structure, called a nested mixture of experts, to represent and learn complex robotic systems and behaviors including hybrid interactions such as contact changes. Our network structure provides a tool for incorporating prior knowledge about the robot's interactions with the environment as well as kinematic and dynamic properties. We devise a loss function to train the local experts to be competitive or cooperative, which enables our NMOE to be trained quickly, to generalize to unseen data from limited data, and to optimize bias-variance trade-off. In various continuous control domains, we evaluate our NMOE with related baseline models both in a supervised learning and an MBRL setup.

% Acknowledgments---Will not appear in anonymized version
\acks{The authors would like to thank the members of the Human Centered Robotics Laboratory at The University of Texas at Austin for their great help and support. This work was supported by the Office of Naval Research, ONR Grant \#N000141512507.}

\bibliography{2021_l4dc}

%% arxiv
\newpage
\appendix
\section{Hyperparameters}
\label{sec:app_a}
\subsection{Dynamics Model Architectures}
Here, we discuss the network structure of our NMOE as well as other baseline models for each experimental domain.

\subsubsection{Reacher}

\textbf{NMOE}: For the NMOE model of Reacher, we composed $f_1^{(1)}$ with $\text{FC}(32,\text{elu}) \rightarrow \text{FC}(32,\cdot)$, where $\text{FC}(a, b)$ represents a fully-connected layer with $a$ nodes and activation function, $b$. The gating network $g^{(1)}$ has an architecture of $\text{FC}(16,\text{softmax})$. The trainable variables include the weights and bias in the fully-connected layers as well as the mass and length parameters in the equation of motions.
\\
\textbf{BB}: For the single black-box model, we represented the dynamics with $\text{FC}(32,\text{elu}) \rightarrow \text{FC}(32,\cdot)$. The trainable variables include the weights and bias in the fully-connected layers.
\\
\textbf{MBB} and \textbf{MWB}: In the case of Reacher, there is one single contact mode. Therefore, the MBB model is identical to the BB model and we omitted it in comparison. Similarly, the MWB model is composed of a single white-box model whose trainable variables only include the mass and length parameters in the equation of motions.

\subsubsection{CartWithElasticWall}

\textbf{NMOE}: For the NMOE model of Eq.~\eqref{eq:cart}, we used $\text{FC}(32,\text{elu}) \rightarrow \text{FC}(32,\cdot)$ to represent $f_1^{(1)}$ and $f_1^{(2)}$. The gating network $g^{(1)}$, $g^{(2)}$, and $h$ have an architecture of $\text{FC}(16, \text{softmax})$, respectively. The trainable variables include the weights and bias in the fully-connected layers as well as $m$, $k$, $b$, and $\theta$ in the equation of motions.
\\
\textbf{BB}: For the single black-box model, we represented the dynamics with $\text{FC}(32,\text{elu}) \rightarrow \text{FC}(32,\cdot)$. The trainable variables include the weights and bias in the fully-connected layers.
\\
\textbf{MBB}: For the mixture of black-box model, we composed $f_1^{(1)}$ and $f_1^{(2)}$ with $\text{FC}(32,\text{elu}) \rightarrow \text{FC}(32,\cdot)$, respectively. The gating network $h$ has an architecture of $\text{FC}(16, \text{softmax})$. The trainable variables include the weights and bias in the fully-connected layers.
\\
\textbf{MWB}: For the mixture of white-box model, the gating network $h$ is $\text{FC}(16, \text{softmax})$. The trainable variables include the weights and bias in the gating network as well as $m$, $k$, $b$, and $\theta$ in the equation of motions.

\subsubsection{Hopper and HalfCheetah}

\textbf{NMOE}: For the NMOE model of the Hopper and HalfCheetah, we composed $f_1^{(1)}$, $f_1^{(2)}$, $f_1^{(3)}$ and $f_1^{(4)}$ with  $\text{FC}(32,\text{elu}) \rightarrow \text{FC}(32,\cdot)$, respectively. The gating network $g^{(1)}$, $g^{(2)}$, $g^{(3)}$, $g^{(4)}$ and $h$ have an architecture of $\text{FC}(16, \text{softmax})$, respectively. The trainable variables include the weights and bias in the fully-connected layers as well as the mass parameters in the equation of motions.
\\
\textbf{BB}: For the single black-box model, we used $\text{FC}(256,\text{elu}) \rightarrow \text{FC}(256,\text{elu}) \rightarrow \text{FC}(256,\cdot)$ to represent dynamics model. Note that we used a deeper and wider network compared to the NMOE model. The trainable variables include the weights and bias in the fully-connected layers. 
\\
\textbf{MBB}: For the mixture of black-box model, we used $\text{FC}(128,\text{elu}) \rightarrow \text{FC}(128,\cdot)$ for all black-box models, $f_1^{(1)}$, $f_1^{(2)}$, $f_1^{(3)}$ and $f_1^{(4)}$. The gating network $h$ has an architecture of $\text{FC}(16, \text{softmax})$. The trainable variables include the weights and bias in the fully-connected layers.
\\
\textbf{MWB}: For the mixture of white-box model, the gating network $h$ is $\text{FC}(16, \text{softmax})$. The trainable variables include the weights and bias in the gating network as well as the mass parameters in the equation of motions.

\subsection{Model-based Control}
Hyperparameters used for model-based control are summarized in Table~\ref{tab:mppi_param}. Here, the time horizon, \# of samples, temperature, and system noise correspond to the variable $T$, $K$, $\lambda$, and $\Sigma$ in the work of \cite{mppi}.

\begin{table}[h!]
\centering
\begin{tabular}{c c c c c} 
 \hline
 & Reacher & CartWithElasticWall & Hopper & HalfCheetah \\ 
 \hline\hline
 Time horizon & 64 & 64 & 128 & 128 \\
 \# samples & 64 & 64 & 128 & 128  \\
 Temperature & 0.01 & 0.01 & 0.01 & 0.01  \\
 System noise & 0.15 & 0.15 & 0.1 & 0.1 \\
 \hline
\end{tabular}
\caption{Hyperparameters used in the model predictive path integral control.}
\label{tab:mppi_param}
\end{table}

\section{White-box Modeling for Hopper and HalfCheetah}
\label{sec:app_b}
Let us start with the dynamics equation of the robot contacting the ground as
\begin{align}
    \label{eq:ap_1}
    A\ddot{q} + b + g = S^{\top}\tau + J_c^{\top} F_c, \\
    \label{eq:ap_2}
    J_c \ddot{q} + \dot{J_c} \dot{q} = 0.
\end{align}
Here, $q \in \mathbb{R}^{n_q}$, $\tau \in \mathbb{R}^{n_\tau}$, $A \in \mathbb{R}^{n_q \times n_q}$, $b \in \mathbb{R}^{n_q}$, and $g \in \mathbb{R}^{n_q}$ are the joint positions, joint torques, mass matrix, coriolis vector and the gravity vector in generalized coordinate, where $n_q$ and $n_\tau$ are the number of joints and actuators. $S^{\top}$ is the selection matrix that maps joint space to actuator space. $J_c \in \mathbb{R}^{n_c \times n_q}$ represents contact space jacobian and $F_c$ denotes reaction force from the environment, where $n_c$ is the dimension of contact space (e.g., $3$ for a point contact). Eq.~\eqref{eq:ap_1} is the Euler-Lagrangian Equation and Eq.~\eqref{eq:ap_2} describes the unilateral contact constraint. Left multiplying $J_c A^{-1}$ to Eq.~\eqref{eq:ap_1} yields
\begin{equation}
    \label{eq:ap_3}
    J_c \ddot{q} + J_c A^{-1} (b+g) = J_c A^{-1}S^{\top}\tau + J_c A^{-1} J_c^{\top} F_c.
\end{equation}
Substituting Eq.~\eqref{eq:ap_2} into Eq.~\eqref{eq:ap_3} results in
\begin{equation}
    \label{eq:ap_4}
    -\dot{J}_c \dot{q} + J_c A^{-1} (b+g) - J_c A^{-1} S^{\top} \tau = J_c A^{-1} J_c^{\top} F_c.
\end{equation}
Let us define $\Lambda^{-1}_c \coloneqq J_c A^{-1} J_c^{\top}$ and multiply $\Lambda_c$ by Eq.~\eqref{eq:ap_4} yields
\begin{equation}
    \label{eq:ap_5}
    -\Lambda_c\dot{J}_c \dot{q} + \Lambda_c J_c A^{-1} (b+g) - \Lambda_c J_c A^{-1} S^{\top} \tau = F_c.
\end{equation}
We further define $\overline{J_c} \coloneqq A^{-1} J_c^{\top} \Lambda_c$ and rewrite Eq.~\eqref{eq:ap_5} with
\begin{equation}
    -\Lambda_c\dot{J}_c \dot{q} + \overline{J_c}^{\top} (b+g) - \overline{J_c}^{\top} S^{\top} \tau = F_c.
\end{equation}
Substituting $F_c$ into Eq.~\eqref{eq:ap_3} results in
\begin{equation}
    A\ddot{q} + (I-J_c\overline{J_c}^{\top})(b+g)+J_c^{\top}\Lambda_c\dot{J}_c\dot{q} = (I-J_c\overline{J_c}^{\top}) S^{\top} \tau.
\end{equation}
We define $N_c \coloneqq I - \overline{J_c}J_c^{\top}$ and solve for $\ddot{q}$ as
\begin{equation}
    \ddot{q} = A^{-1}(N_{c}^{\top}(S^{\top}\tau_k-b-g)-J_{c}^{\top}\Lambda_{c}\dot{J}_{c}\dot{q}).
\end{equation}
Finally, the white-box modeling in the manuscripts are achieved by the Euler integration.

\section{Reward Functions in Model-based Control}
\label{sec:app_c}
\begin{figure*}[t]
    \centering
    \includegraphics[width=0.6\linewidth]{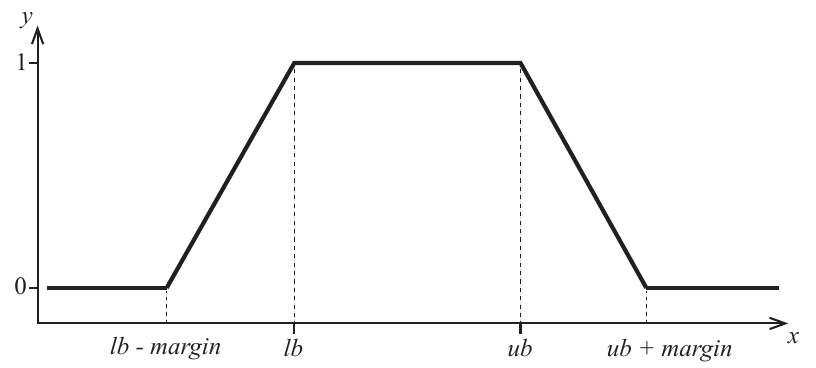}
    \caption{$y=\texttt{tolerance}(x,(lb,ub),margin)$}
    \label{fig:tolerance}
\end{figure*}

Here, we describe the reward function for each experimental domain we used in the MBRL setup. In the experiment, we normalize the reward signal from using a function \texttt{tolerance}$: \mathbb{R} \mapsto \mathbb{R}$ which is introduced in \cite{dm_control} and illustrated in Figure~\ref{fig:tolerance}. The function returns $1$ when $lb \leq x \leq ub$. The output decreases linearly with the $margin$ distance from the interval.

\subsection{Reacher}
The two-link manipulator receives a reward signal $1$ when the end-effector reaches near the target location. The reward function is defined as follow:
\begin{equation}
    r = \texttt{tolerance}\left(\vert \vert  x_{\text{ee}}^d - x_{\text{ee}} \vert \vert_2^2, (0, 0.03), 0\right)
\end{equation}
where $x_{\text{ee}}^d$ and $x_{\text{ee}}$ denote the desired and the actual end-effector position of the manipulator.

\subsection{CartWithElasticWall}
The cart receives a reward signal $1$ when the cart reaches near the target location. The reward function is defined as follow:
\begin{equation}
    r = \texttt{tolerance}(x, (x_d, \infty), 0)
\end{equation}
where $x$ is the cart position and $x_d$ is the target height.

\subsection{Hopper}
The hopper agent receives a reward signal when it maintains the torso height $q_1$ and to forward velocity $\dot{q}_0$ close to the desired values. The reward function is defined as follow:
\begin{equation}
    r = \texttt{tolerance}(q_1, (0.7, \infty),0) \cdot \texttt{tolerance}(\dot{q}_0, (2, \infty),2).
\end{equation}

\subsection{HalfCheetah}
The HalfCheetah receives a reward signal when it maintains its body orientation $q_2$ and to forward velocity $\dot{q}_0$ close to the desired values. The reward function is defined as follow:
\begin{equation}
    r = \texttt{tolerance}(q_2, (-0.35, 0.35),0) \cdot \texttt{tolerance}(\dot{q}_{0}, (10, \infty), 10).
\end{equation}

% \bibliography{2021_l4dc}

\end{document}